\documentclass[11pt,a4paper]{article}
\usepackage[utf8]{inputenc}
\usepackage[british]{babel}
\usepackage[margin=1in]{geometry}
\usepackage{hyperref}
\usepackage{graphicx}
\usepackage{amsmath}
\usepackage{amssymb}
\usepackage{booktabs}
\usepackage{subcaption}
\usepackage{csquotes}
\usepackage{enumitem}
\usepackage{tikz}
\usepackage{xcolor}
\usepackage[numbers,sort&compress]{natbib}

\title{\textbf{CALF: Communication-Aware Learning Framework for Distributed Reinforcement Learning}}

\author{%
Carlos Purves \\
University of Cambridge \\
\texttt{cp614@cam.ac.uk}
\and
Pietro Li\`o \\
University of Cambridge \\
\texttt{pl219@cam.ac.uk}
}

\date{}

\begin{document}

\maketitle

\begin{abstract}
Distributed reinforcement learning policies face network delays, jitter, and packet loss when deployed across edge devices and cloud servers. Standard RL training assumes zero-latency interaction, causing severe performance degradation under realistic network conditions. We introduce CALF (Communication-Aware Learning Framework), which trains policies under realistic network models during simulation. Systematic experiments demonstrate that network-aware training substantially reduces deployment performance gaps compared to network-agnostic baselines. Distributed policy deployments across heterogeneous hardware validate that explicitly modelling communication constraints during training enables robust real-world execution. These findings establish network conditions as a major axis of sim-to-real transfer for Wi-Fi-like distributed deployments, complementing physics and visual domain randomisation.
\end{abstract}

\section{Introduction}

Reinforcement learning policies deployed across distributed hardware, whether environments run on edge devices and policies execute on remote servers or control is partitioned across heterogeneous processors, face network delays, jitter, and packet loss in the agent-environment control loop. Yet mainstream RL training assumes synchronous, zero-latency interaction. Standard benchmarks (ALE~\citep{Bellemare2013ALE}, DeepMind Control Suite~\citep{Tassa2018DMC}, OpenAI Gym~\citep{Brockman2016Gym}) presuppose instant observation delivery and immediate action effects, while distributed training systems (IMPALA~\citep{Espeholt2018IMPALA}, SEED RL~\citep{Espeholt2019SEEDRL}) optimise worker-learner communication but treat agent-environment communication as an implementation detail. Hierarchical policies, which decompose control into modular units for specialisation and skill reuse~\citep{Sutton1999Options, Parr1998HAM, Dietterich2000MAXQ, Bacon2017OptionCritic, vezhnevets2017feudal, Nachum2018HIRO}, face additional challenges when distributed, as inter-unit communication introduces network delays on top of agent-environment delays.

\begin{figure}[t]
    \centering
    \includegraphics[width=0.9\linewidth]{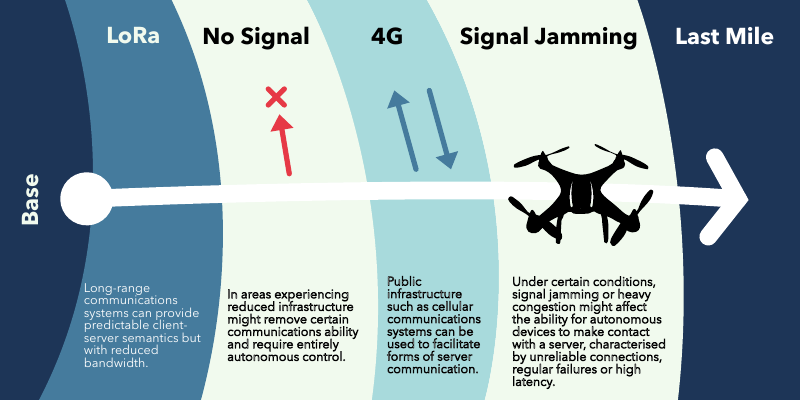}
    \caption{Real-world distributed systems employ hybrid communication strategies to maintain operation under varying network conditions. Unmanned aerial vehicles, for instance, choose between multiple communication channels (satellite, radio frequency, optical tether) based on signal quality and operational constraints, defaulting to autonomous operation when no reliable connection exists. This illustrates the challenge CALF addresses: policies must function across heterogeneous network conditions rather than assuming perfect connectivity. We focus on LAN-like scenarios (Wi-Fi, Ethernet); WAN/adversarial scenarios motivate the need for configurable impairments but are not evaluated here.}
    \label{fig:systems:uav_communication}
\end{figure}

In deployment, these assumptions fail. Observations arrive late or out-of-order; actions are delayed or dropped; jitter creates unpredictable timing. A policy that perfectly balances an inverted pendulum in simulation may fail with 100~ms Wi-Fi latency, even with perfect physics modelling. Sim-to-real transfer has made substantial progress addressing physics and visual mismatch through domain randomisation~\citep{Tobin2017DomainRandomization, Peng2018SimToReal}, but network-induced mismatch receives minimal attention despite evidence that actuator dynamics, including timing and response behaviour, are a dominant factor in the sim-to-real gap~\citep{Hwangbo2019Learning}. Network conditions constitute an orthogonal axis of the reality gap that must be addressed systematically.

This paper addresses three research questions:

\textbf{RQ1 (Network Impact):} How severely do realistic network conditions, including latency, jitter, and packet loss, degrade the performance of policies when trained in idealised, synchronous simulations but deployed over real networks with distributed components?

\textbf{RQ2 (Network-Aware Training):} Can training policies under realistic network conditions during simulation (``network-aware training'') close this performance gap? Which network phenomena (latency versus jitter versus loss) are most critical to model?

\textbf{RQ3 (Infrastructure):} What systems infrastructure is needed to enable reproducible, scalable deployment of distributed RL across heterogeneous edge devices and real networks?

We introduce CALF (Communication-Aware Learning Framework), infrastructure for deploying and training distributed RL policies across heterogeneous hardware. CALF implements policy units as networked services and transparently injects configurable network impairments (latency, jitter, loss, bandwidth limits) via NetworkShim middleware, enabling the same policy code to run from pure simulation to distributed edge-cloud hardware. Systematic experiments on CartPole and MiniGrid demonstrate that baseline policies suffer severe degradation (40--80\%) when deployed over realistic networks (RQ1), while network-aware training reduces the sim-to-real gap by approximately $4\times$ for CartPole and $3\times$ for MiniGrid (RQ2). Ablations reveal that stochastic jitter and packet loss are more detrimental than constant latency. Distributed policy deployments across Raspberry Pi and Desktop hardware validate CALF's practicality for edge-cloud systems (RQ3).

The remainder positions CALF relative to prior work (§2), describes the framework architecture (§3), methodology (§4), and experimental setup (§5), presents results (§6), and discusses implications (§7).

\section{Background and Positioning}
\label{sec:related}

\subsection{Delays, Packet Loss, and What RL Typically Assumes}

Early work extended the Markov Decision Process framework to include action and observation delays. Katsikopoulos \& Engelbrecht~\cite{Katsikopoulos2003DelayedMDP} showed that fixed $k$-step delays can be transformed into an equivalent Markov process by augmenting the state with the last $k$ actions or observations. However, this state augmentation causes exponential state-space growth, rendering planning or learning intractable for large delays. Walsh et al.~\cite{Walsh2007DelayedMDP} proved an exponential lower bound: no algorithm can circumvent this blow-up in the worst case. With stochastic delays, optimal policies must use full history, becoming POMDP-like~\cite{Katsikopoulos2003DelayedMDP}.

Practical RL uses heuristics and algorithmic fixes. Simple strategies include frame stacking (DQN~\cite{Mnih2015DQN} uses the last 4 observations to infer velocity) and action repetition (holding commands for $k$ steps). These work for small, fixed delays but fail with variable delays or packet loss. Schuitema et al.~\cite{Schuitema2010DelayedRL} proposed delay-aware TD learning that updates the \emph{delayed-next-state} rather than immediate next-state, avoiding state augmentation but remaining sub-optimal for variable delays. Bouteiller et al.~\cite{Bouteiller2021DelayRL} introduced Delay-Correcting Actor-Critic (DCAC), which resamples and relabels experience trajectories to correct for random delay distortions. A consistent finding is that unmitigated latency severely degrades performance, but training under delays yields robustness.

Control theory has extensively studied networked control systems (NCS)~\cite{Hespanha2007NCS} where sensors, controllers, and actuators communicate over imperfect networks. Standard compensation strategies include zero-order hold (actuators keep executing previous commands when packets are dropped), time-stamping and sequence numbers (to detect stale or out-of-order packets), Smith predictors (which compensate for known constant delays), and event-triggered control (sending updates only when state error exceeds a threshold). Researchers have derived stability conditions on maximum tolerable delay or dropout rate using Lyapunov methods for linear systems~\cite{Hespanha2007NCS}. However, NCS analysis typically applies to linear controllers with analytical models. Deep RL policies are black-box nonlinear functions for which no similar guarantees exist, and systematic application of NCS insights to deep RL remains limited.

\subsection{Sim-to-Real Transfer and Distributed RL: The Missing Network Axis}

Sim-to-real RL focuses overwhelmingly on physics and visual domain randomisation, with minimal attention to network-induced mismatch. Domain randomisation~\cite{Tobin2017DomainRandomization} randomises simulator properties so the real world appears as another random variant, enabling zero-shot transfer for manipulation~\cite{Tobin2017DomainRandomization,OpenAI2019Dactyl} and locomotion~\cite{Tan2018SimToRealQuadruped}. Alternative approaches include system identification~\cite{Chebotar2019DART}, which iteratively fits simulator parameters to real data, and online adaptation~\cite{Peng2018SimToReal}, which uses RNNs to adapt to different dynamics. Hwangbo et al.~\cite{Hwangbo2019Learning} demonstrated spectacular legged locomotion results by carefully modelling actuator dynamics, finding that accurate modelling of the Series Elastic Actuator behaviour was a dominant factor in closing the sim-to-real gap for their ANYmal quadruped. However, they treated this as a robot-specific engineering fix rather than general methodology.

These works randomise physical dynamics and visual observations but \textbf{assume perfect timing}: either the policy and environment are co-located, or network effects are negligible. Network conditions represent an \textbf{independent axis of variation}, orthogonal to physics and vision. A policy trained with perfect timing in simulation may fail on a real system with 100~ms lag, even if the physics are modelled perfectly. Some sim-to-real practices implicitly touch on network effects (Gaussian noise partially mimics jitter, frame skip reduces effective control frequency), but these are incidental rather than systematic strategies to address network domain shift.

Large-scale distributed RL frameworks similarly treat network communication as a cost to minimise, not an object of study. IMPALA~\cite{Espeholt2018IMPALA} separates actors and learners with V-trace off-policy correction to handle policy lag; SEED RL~\cite{Espeholt2019SEEDRL} minimises network overhead with fast transport protocols; Sample Factory~\cite{Petrenko2020SampleFactory} avoids network communication entirely. These systems address network lag between actor and learner (training infrastructure), whereas CALF addresses network lag between agent and environment (control loop). Frameworks like Gym-Gazebo~\cite{Zamora2016GymGazebo} and Isaac Gym~\cite{Makoviychuk2021IsaacGym} enable training in simulation and deployment on robots, but assume local or fast LAN connections where network delay is incidental, not controlled or measured.

\subsection{Edge, Multi-Agent, and Other Network-Aware ML Contexts}

Edge machine learning research focuses primarily on computation and energy constraints, with less attention to communication. Techniques include model compression, quantisation, and distillation to fit policies in limited memory. However, complex vision-based policies will not fit on tiny embedded devices, necessitating splitting or offloading. Neurosurgeon~\cite{Kang2017Neurosurgeon} automatically partitions deep neural networks between edge devices and cloud (convolutional layers on edge, fully connected layers on server), achieving $3\times$ lower latency and energy compared to all-cloud or all-device execution.

Multi-agent RL studies how agents learn to communicate under bandwidth limits or delays. Work on emergent communication includes learned continuous communication protocols~\cite{Mao2020MARL} and communication minimisation via information-theoretic regularisation~\cite{Wang2020CommRL}. However, MARL focuses on agent-to-agent delays, while agent-to-environment delays in single-agent control RL remain less explored. Federated RL addresses network issues in parameter aggregation (training infrastructure), not agent-environment interaction loops (control execution).

Hierarchical RL decomposes behaviour into subskills. The Options framework~\cite{Sutton1999Options} introduced temporally extended actions that execute for multiple steps, enabling higher-level decision-making at slower timescales. If an option runs autonomously for 10 steps, the high-level policy only needs to communicate every 10 steps, making it naturally more robust to moderate network latency. Modern variants include Option-Critic~\cite{Bacon2017OptionCritic}, FeUdal Networks~\cite{vezhnevets2017feudal}, and HIRO~\cite{Nachum2018HIRO}. However, existing hierarchical RL research typically assumes components are co-located, while distributed deployment across heterogeneous hardware introduces network communication as a first-order constraint. Mature network emulation tools exist (Linux \texttt{tc netem}, Mininet~\cite{Lantz2010Mininet}, Mahimahi~\cite{Netravali2015Mahimahi}) but are not integrated into RL training loops.

In contrast to prior work that modifies algorithms (delay-aware Q-learning, DCAC) or optimises training infrastructure (IMPALA, SEED RL), CALF modifies the environment to expose realistic network behaviour, an approach that is algorithm-agnostic and extends naturally to heterogeneous edge deployment. By implementing policy units as networked services and transparently injecting network impairments, CALF complements existing domain randomisation practices by adding network parameters to the randomisation distribution. Together, these strands of work suggest two requirements for progress: training must experience the same communication pathologies as deployment, and the infrastructure must allow controlled, reproducible manipulation of latency, jitter, and loss across real hardware. CALF is designed to meet both.

\section{CALF: A Framework for Network-Aware Reinforcement Learning}
\label{sec:framework}

CALF decomposes RL workloads into networked services, injects realistic network behaviours at specific communication links, and runs the same configuration across deployment modes from pure simulation to real hardware.

\subsection{Design Goals}

CALF enables network-aware RL research by providing: (1) realistic network conditions (configurable latency, jitter, loss distributions; synthetic and trace-based models), (2) deployment parity (same policy code runs from pure simulation to edge hardware), (3) reproducibility (deterministic network seeds, versioned modules, containerisation), and (4) heterogeneous hardware support (edge devices to cloud servers). CALF is algorithm-agnostic infrastructure compatible with any RL library (Table~\ref{tab:design_goals}).

\begin{table}[h]
\centering
\small
\caption{CALF design goals and their role in answering research questions.}
\label{tab:design_goals}
\begin{tabular}{llp{5cm}}
\toprule
\textbf{Goal} & \textbf{Capability} & \textbf{Enables} \\
\midrule
Network Realism & Synthetic + trace-based network models & Controlled ablations (RQ2), realistic training \\
Deployment Parity & Same code across simulation/hardware & Fair comparison of network effects (RQ1) \\
Reproducibility & Deterministic seeds, versioning & Scientific rigour, exact replication \\
Heterogeneity & Edge devices to cloud servers & Realistic distributed settings (RQ3) \\
\bottomrule
\end{tabular}
\end{table}

\subsection{Architecture Overview}

CALF realises policies and environments as networked services communicating via message passing. Unlike traditional RL where \texttt{obs = env.step(action)} is a zero-latency function call, CALF implements \texttt{step()} as network communication, thereby enabling both distributed deployment and controlled injection of network impairments. Agent Services (policy units), Environment Services, and NetworkShim Services (impairment injectors) compose into policy graphs, with NetworkShims sitting on graph edges to introduce realistic latency, jitter, and loss. CALF includes orchestration and optional cross-network routing; our experiments require only service composition plus impairment injection. The architecture's key property is deployment parity: the same service configuration runs locally (zero latency), with simulated networks (NetworkShim impairments), or across real hardware (Wi-Fi/Ethernet). Full infrastructure details are available in the accompanying code release.

\subsection{NetworkShim: The Core Mechanism}
\label{subsec:networkshim}

NetworkShim is CALF's primary mechanism for injecting network impairments into the RL loop. It acts as a transparent middlebox sitting between Environment and Agent, delaying or dropping packets according to configured network models. All messages are timestamped to measure and replay latency distributions ($\text{latency}_{\text{ms}} = t_{\text{receive}} - t_{\text{send}}$); NetworkShim uses this to inject delay and loss while logging realised conditions. For each packet, NetworkShim simulates loss (drop with probability $p_{\text{loss}}$) and delay (sampled from configured distributions: $\max(0, \mathcal{N}(\mu_{\text{latency}}, \sigma_{\text{jitter}}^2))$ for jitter, or a fixed value for constant latency), then forwards the packet after the sampled delay.

NetworkShim supports two families of network model. Synthetic models define parametric distributions matching our evaluation conditions: \emph{Ethernet-clean} (2~ms $\pm$ 0.5~ms, 0\% loss), \emph{Wi-Fi-normal} (30~ms $\pm$ 10~ms, 2\% loss), and \emph{Wi-Fi-degraded} (80~ms $\pm$ 40~ms, 10\% loss), with latency sampled from normal distributions clipped at zero and loss drawn from a Bernoulli($p$). Trace-based models instead enable replay of recorded conditions: a LatencyTracer service calculates actual latency from packet timestamps during Real-Wi-Fi evaluation and logs traces, which NetworkShim can then replay during training by sampling delays from the empirical distribution. This allows policies trained on a synthetic Wi-Fi-normal model to be refined using real Wi-Fi traces, or enables controlled experiments comparing conditions such as ``Real-Wi-Fi-Home'' versus ``Real-Wi-Fi-Campus''.

Critically, Environment and Agent are unaware of NetworkShim's existence; they simply experience delayed messages. This transparency enables network-aware training without modifying RL algorithms.

\subsection{Progressive Deployment Modes}
\label{subsec:deployment}

\begin{figure}[t]
    \centering
    \includegraphics[width=0.95\linewidth]{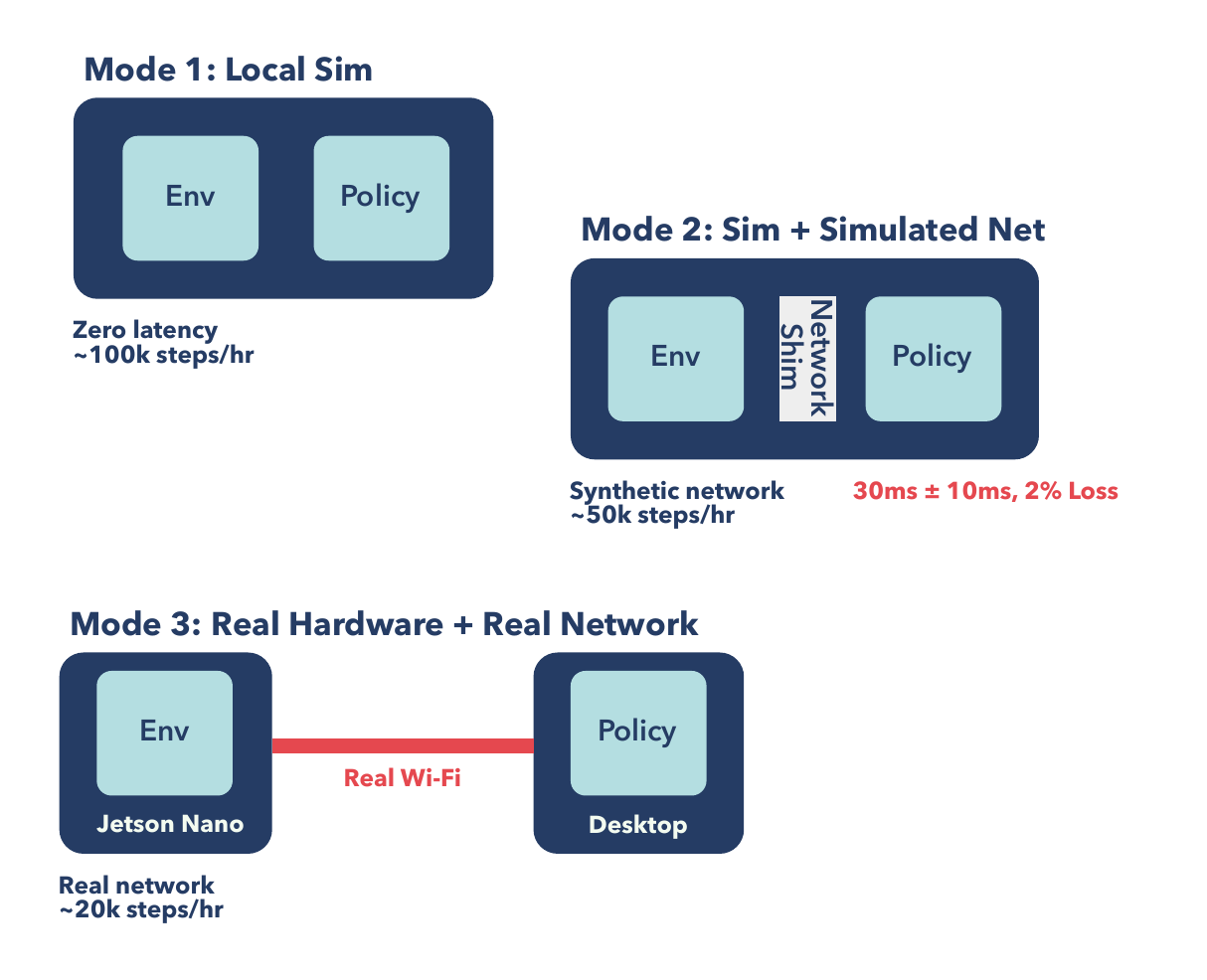}
    \caption{CALF's three progressive deployment modes enable incremental validation from pure simulation to distributed deployment. Mode~1 (Local Sim) provides a zero-latency baseline for rapid development with environment and policy co-located. Mode~2 (Sim + Simulated Network) introduces NetworkShim services that inject realistic latency, jitter, and packet loss for network-aware training. Mode~3 (Edge Sim) validates distributed deployment on real hardware (Raspberry Pi for environment, Desktop for policy) communicating over real Wi-Fi/Ethernet networks. This progressive approach ensures that network-aware policies trained in Mode~2 transfer successfully to distributed edge deployment in Mode~3, addressing the network axis of the sim-to-real gap.}
    \label{fig:deployment_modes}
\end{figure}

A key CALF feature is that the same policy and environment code run across a continuum of deployment scenarios (Figure~\ref{fig:deployment_modes}). In Mode~1 (Local Sim), the environment and policy reside in the same process and communicate via direct function calls with no network overhead, providing a zero-latency baseline for fast prototyping and comparison (RQ1) at approximately 100K steps/hour for CartPole on the Desktop. In Mode~2 (Sim + Simulated Network), the environment and policy run as separate services with a CALF NetworkShim between them, injecting impairments drawn from a synthetic network model (e.g., Wi-Fi-normal: 30~ms $\pm$ 10~ms, 2\% loss); this mode is used for network-aware training (RQ2) and achieves approximately 50K steps/hour, slower owing to injected delays. In Mode~3 (Edge Sim), the environment service runs on a Raspberry Pi (or other edge device such as a Jetson) while the policy service runs on the Desktop, communicating over a real network (Ethernet or Wi-Fi); this mode is used for hardware validation and for measuring real network distributions, achieving approximately 20K steps/hour as throughput is limited by the network and Pi CPU. Together, the three modes progressively de-risk deployment: rapid iteration in Mode~1, exposure to realistic delays in Mode~2, and detection of hardware-specific issues in Mode~3.

\subsection{Summary}

This section has described CALF's three-layer architecture enabling distributed deployment of policy graphs, NetworkShim middleware transparently injecting network impairments on graph edges, and progressive deployment modes ensuring deployment parity from simulation to real hardware. CALF ensures reproducibility through versioned modules, containerisation, and deterministic network seeds (NetworkShim uses fixed RNG seeds for reproducible delay sequences), and all experiments in this paper are reproducible with the accompanying code release. Of the three modes, Mode~2 is the key experimental lever: it isolates the network axis by introducing impairments while holding the environment and policy code fixed. We use Mode~2 to train under controlled stochastic network models, then evaluate transfer to Mode~3 real networks.

\section{Network-Aware Training Methodology}
\label{sec:methodology}

This section describes our RL training protocol and experimental methodology for answering RQ1 (how severely do network conditions degrade performance of distributed policy graphs?) and RQ2 (does network-aware training enable successful policy graph deployment?).

\subsection{Problem Formulation: Delayed MDPs}

In deployment, observations and actions are delayed or dropped, making the policy effectively partially observable: the agent must act on stale information and infer the current state from observation history. We address this via frame stacking (CartPole) or recurrence (MiniGrid) to maintain temporal context. See Section~\ref{sec:related} for formal delayed MDP background.

\subsection{Training Regimes: Comparing Network-Awareness}

Table~\ref{tab:training_regimes} summarises our three training regimes. We train policies under each regime and evaluate all policies on all deployment modes, enabling systematic comparison of network-agnostic versus network-aware training.

\begin{table}[h]
\centering
\caption{Training regimes for network-aware evaluation.}
\label{tab:training_regimes}
\small
\begin{tabular}{lccc}
\toprule
\textbf{Regime} & \textbf{Mode} & \textbf{Impairments} & \textbf{Purpose} \\
\midrule
Baseline & Mode~1 (local) & None & Traditional RL (no awareness) \\
Delay-Only & Mode~2 & 50ms fixed latency & Constant-delay awareness \\
Full Net-Aware & Mode~2 & Latency + jitter + loss & Full stochastic model \\
\bottomrule
\end{tabular}
\end{table}

Network statistics for full network-aware training are measured during pilot runs using LatencyTracer and modelled as $\mathcal{N}(\mu, \sigma^2)$ for latency (Wi-Fi-normal: mean 30ms, jitter 10ms), with 2\% packet loss.

Regimes differ only in the communication channel; we hold everything else fixed (environment, architecture, training procedure). To prevent the comparison from being confounded by observability, we provide temporal context via memory/stacking, enabling each policy to infer current state from delayed observations regardless of training regime.

\subsection{RL Algorithm: PPO}

We use Proximal Policy Optimization~\cite{Schulman2017PPO} via Stable-Baselines3~\cite{Raffin2021SB3} as a representative modern RL algorithm. CALF is algorithm-agnostic; others (SAC, DQN) could be trained with identical network configurations. For policy graphs, each unit can employ different algorithms suited to its role. Hyperparameters follow standard settings (see the code release for details).

\subsection{State Representation for Delay Robustness}

Policy units must infer current state from delayed observations. We employ three strategies depending on task characteristics (Table~\ref{tab:strategies}): frame stacking to infer velocities from multiple snapshots, recurrent policies (LSTM) to maintain belief state over variable delays, and action history to track in-flight actions. CartPole uses frame stacking ($k = d+1$ frames for delay $d$); MiniGrid uses LSTM (necessary for partial observability combined with delays).

\begin{table}[h]
\centering
\small
\caption{State representation strategies for handling observation delays.}
\label{tab:strategies}
\begin{tabular}{lll}
\toprule
\textbf{Task} & \textbf{Representation} & \textbf{Rationale} \\
\midrule
CartPole & Frame stack $k{=}d{+}1$ & Infer velocities under delay \\
MiniGrid & LSTM & Variable delay + partial observability \\
Ablation & + Action history & Track in-flight actions \\
\bottomrule
\end{tabular}
\end{table}

\subsection{Evaluation Protocol}

For each trained policy (each seed, each training regime), we evaluate on five deployment modes grouped into two categories: \textbf{Sim} (Sim-Clean [Mode 1, local], Sim+Network [Mode 2, Wi-Fi-normal model]) and \textbf{Real} (Real-Ethernet, Real-Wi-Fi-Normal, Real-Wi-Fi-Degraded [Mode 3 variants, Pi environment + Desktop policy]). Per mode, we run 50 episodes and record episodic return, success rate, and timestamped end-to-end latency.

For statistical rigour, we use 10 random seeds per training regime, report mean $\pm$ standard deviation across seeds, use paired $t$-tests to compare full network-aware versus baseline in Real-Wi-Fi-Degraded, and use significance level $\alpha = 0.05$.

\section{Experimental Setup}
\label{sec:setup}

This section specifies environments, agents, hardware platforms, and evaluation metrics for complete reproducibility (G3).

\subsection{Environments}

We select environments that span diverse timing sensitivities, are well-known benchmarks familiar to the community, and remain tractable on modest hardware.

CartPole-v1 is a classic inverted-pendulum task in which the agent must balance a pole on a movable cart. The state is 4-dimensional (cart position, cart velocity, pole angle, pole angular velocity), the action space is discrete $\{$left, right$\}$, and the episode terminates when $|x| > 2.4$, $|\theta| > 12^\circ$, or 500 steps are reached, with a reward of +1 per step survived (maximum 500). CartPole is highly timing-sensitive: its unstable dynamics require fast reactions, so delays easily destabilise control, and a 100~ms delay can cut survival time in half. The failure mode is unambiguous (the pole falls), making results straightforward to interpret.

MiniGrid DoorKey-8x8 is a gridworld navigation task with a locked door: the agent must find a key, unlock the door, and reach the goal. The observation is a $7 \times 7$ egocentric view (partial observability), the action space is discrete $\{$turn left, turn right, move forward, pick up, toggle$\}$, and the episode terminates upon reaching the goal (success) or after 1000 steps (failure), with a reward of +1 for success and $-0.01$ per step. MiniGrid is less sensitive to millisecond timing than CartPole but is still affected by delays, as agents may overshoot turns or collide with walls. Its subgoal structure (get key, open door, reach goal) provides a natural hierarchy and demonstrates the potential for decomposition into separate ``key policy'' and ``goal policy'' units.

\subsection{Agent Architectures}

For the primary experiments we use flat policies trained with PPO over 10 random seeds per regime. The CartPole policy is a two-hidden-layer multi-layer perceptron (64 units per layer, ReLU activations) producing 2-dimensional action logits from a 4-dimensional observation (or $4 \times k$ when frame-stacked). The MiniGrid policy is a convolutional network that processes the $7 \times 7 \times 3$ observation through two convolutional layers (16 and 32 filters, both $3 \times 3$), followed by flattening, an LSTM with 128 units, a fully connected layer of 128 units, and 5-dimensional action logits.

To demonstrate CALF's support for distributed hierarchical policies, we additionally test two-level policy graph deployments. The CartPole graph comprises angle-stabiliser and recentring units with rule-based switching ($|\theta| > 5^\circ$); the MiniGrid graph comprises key-seeking and goal-navigation units with state-based switching (\texttt{has\_key}). Units are trained independently in local simulation with task-specific rewards and then deployed across Pi (edge) and Desktop (cloud) with NetworkShim on inter-unit communication. This demonstrates cross-device execution capability rather than optimal hierarchical decomposition.

\subsection{Hardware and Network Conditions}

Experiments use a Desktop (Intel i7, 32~GB RAM) as policy host and a Raspberry Pi 4 Model B (ARM Cortex-A72, 4~GB RAM) as environment host, representing a heterogeneous edge-cloud deployment. Full specifications are provided in the code release.

Three network configurations span the range from ideal to degraded conditions. Ethernet-Clean connects the Desktop and Pi via a physical Ethernet cable, yielding a mean latency of 2~ms, jitter of 0.5~ms, 0.0\% loss, and 1~Gbps link capacity. Wi-Fi-Normal places both devices on the same 802.11ac (5~GHz) Wi-Fi network, producing a mean latency of 30~ms, jitter of 10~ms, 2\% loss, and approximately 50~Mbps measured throughput. Wi-Fi-Degraded augments Wi-Fi-Normal with additional controlled delay, jitter, and loss injected via Linux \texttt{netem} to simulate a congested network, resulting in a mean latency of 80~ms, jitter of 40~ms, and 10\% loss (exact configuration in the code release). All network statistics are measured using LatencyTracer during pilot runs and verified across 1000 packet samples; distributions are logged for reproducibility.

\subsection{Evaluation Metrics}

We report episodic return/success rate (task performance) and sim-to-real gap (percentage performance drop from Sim-Clean to Real-Wi-Fi-Degraded); latency/loss are measured to validate network conditions. All results use 10 random seeds per regime; we report mean $\pm$ standard deviation and use paired $t$-tests for baseline versus full network-aware comparisons ($\alpha = 0.05$). Complete experimental code, network configurations, and evaluation logs will be released.

\section{Results}
\label{sec:results}

Network-aware training substantially improves deployment performance over realistic networks. Ablations reveal that stochastic phenomena (jitter, loss) require explicit modelling, and distributed policy-graph deployments demonstrate cross-device execution capability.

\subsection{Network-Aware Training Improves Real Deployment Performance}

\subsubsection{CartPole Results}

Table~\ref{tab:cartpole_results} presents mean episode return (with standard deviation) across 10 random seeds for each training regime and deployment mode. CartPole's maximum achievable return is 500 (pole balanced for the full episode duration), making performance drops immediately interpretable as failure to maintain balance under network constraints.

\begin{table}[h]
\centering
\small
\caption{CartPole: Mean Episode Return ($\pm$ std) over 10 seeds. Each cell reports mean performance across seeds, with each seed evaluated over 50 episodes per deployment mode.}
\label{tab:cartpole_results}
\begin{tabular}{l@{\hspace{0.3cm}}c@{\hspace{0.3cm}}c@{\hspace{0.3cm}}c@{\hspace{0.3cm}}c@{\hspace{0.3cm}}c}
\toprule
Training Regime & Sim-Clean & Sim+Net & Real-Eth & Wi-Fi-N & Wi-Fi-D \\
\midrule
Baseline        & 495 $\pm$ 7   & 310 $\pm$ 48    & 288 $\pm$ 62      & 173 $\pm$ 71         & 92 $\pm$ 54            \\
Delay-Only      & 482 $\pm$ 11  & 468 $\pm$ 16    & 425 $\pm$ 32      & 348 $\pm$ 49         & 218 $\pm$ 58           \\
Full Net-Aware  & 476 $\pm$ 9   & 472 $\pm$ 13    & 458 $\pm$ 22      & 442 $\pm$ 27         & 378 $\pm$ 41           \\
\bottomrule
\end{tabular}
\end{table}

\textbf{Takeaway:} Baseline policies collapse under real Wi-Fi; full network-aware training recovers most of the lost performance. Delay-only training helps but is insufficient, as stochastic jitter and loss are the dominant degraders.

Baseline policies achieve near-optimal simulation performance (495 $\pm$ 7) but collapse under realistic Wi-Fi (92 $\pm$ 54 in Wi-Fi-D, an 81.4\% drop). Network-aware training substantially closes this gap: Wi-Fi-D performance reaches 378 $\pm$ 41 (20.6\% drop from Sim-Clean), a \textbf{3.95$\times$ reduction in deployment gap} ($t(9) = 12.7$, $p < 0.001$, $d = 2.31$). Delay-only training provides intermediate robustness (218 $\pm$ 58), demonstrating that constant delays alone are insufficient and that stochastic jitter and packet loss require explicit modelling. Section~\ref{subsec:ablation} isolates the contributions of latency, jitter, and loss, confirming that variability and packet dropout are the dominant degraders. Real-Ethernet performance (458 $\pm$ 22) is similar to Sim+Network (472 $\pm$ 13) in this task.

\begin{figure}[h]
\centering
\includegraphics[width=0.95\linewidth]{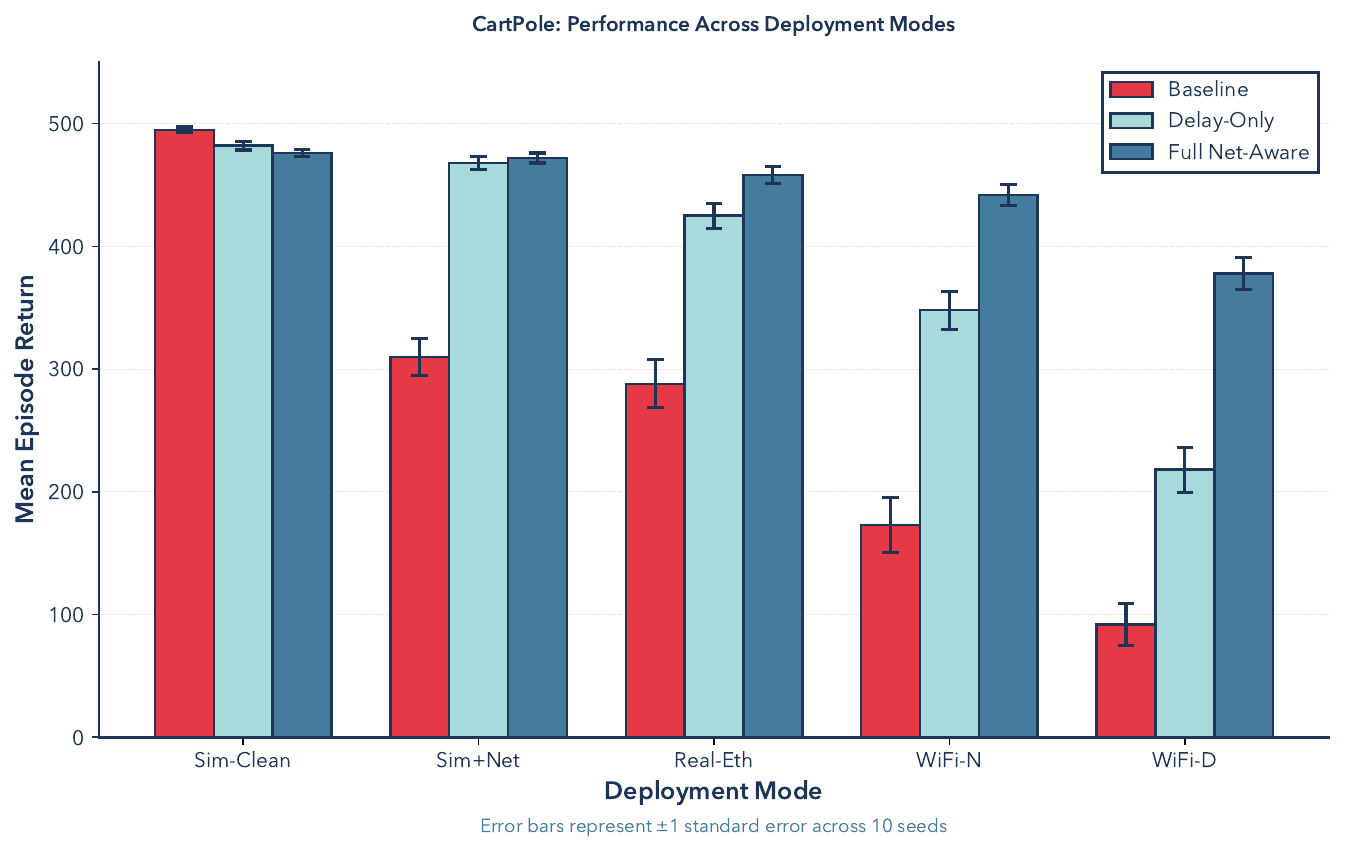}
\caption{CartPole: Performance comparison across deployment modes for each training regime. Full network-aware training maintains robust performance under real network conditions, while baseline training exhibits severe degradation. Delay-only training provides partial robustness, demonstrating the necessity of modelling jitter and packet loss in addition to latency. Error bars represent $\pm$1 std across 10 seeds.}
\label{fig:cartpole_bar}
\end{figure}

\subsubsection{MiniGrid Results}

\begin{table}[h]
\centering
\small
\caption{MiniGrid: Success Rate (\% $\pm$ std) over 10 seeds. Each cell reports percentage of episodes successfully reaching the goal, with each seed evaluated over 50 episodes per deployment mode.}
\label{tab:minigrid_results}
\begin{tabular}{l@{\hspace{0.3cm}}c@{\hspace{0.3cm}}c@{\hspace{0.3cm}}c@{\hspace{0.3cm}}c@{\hspace{0.3cm}}c}
\toprule
Training Regime & Sim-Clean & Sim+Net & Real-Eth & Wi-Fi-N & Wi-Fi-D \\
\midrule
Baseline        & 94 $\pm$ 4    & 76 $\pm$ 9      & 73 $\pm$ 11       & 61 $\pm$ 13          & 44 $\pm$ 16            \\
Delay-Only      & 91 $\pm$ 5    & 87 $\pm$ 6      & 84 $\pm$ 7        & 77 $\pm$ 9           & 64 $\pm$ 11            \\
Full Net-Aware  & 89 $\pm$ 4    & 87 $\pm$ 5      & 85 $\pm$ 6        & 81 $\pm$ 7           & 74 $\pm$ 9             \\
\bottomrule
\end{tabular}
\end{table}

MiniGrid exhibits similar trends with smaller effect sizes. Baseline training achieves 94\% success in Sim-Clean but drops to 44\% in Wi-Fi-D (53.2\% degradation). Network-aware training achieves 74\% (17.0\% degradation), a \textbf{3.13$\times$ reduction in deployment gap} ($t(9) = 8.4$, $p < 0.001$, $d = 1.87$). The smaller magnitude reflects MiniGrid's discrete gridworld dynamics, which are more forgiving than CartPole's unstable continuous control. Delay-only training (64\%) again falls between baseline and full network-aware, confirming that stochastic phenomena require explicit modelling.

\subsection{Impact of Different Network Pathologies}
\label{subsec:ablation}

The preceding results show that delay-only training provides partial robustness but leaves a substantial residual gap: CartPole delay-only achieves 218 versus full network-aware's 378 in Wi-Fi-D, a 160-point difference. We hypothesise that this gap arises from stochastic phenomena, namely variability and dropout, rather than from mean latency alone. To test this, we isolate the individual contributions of latency, jitter, and packet loss.

We conduct an ablation study training CartPole policies under four conditions: latency-only (constant 50~ms delay), stochastic additional delay (at each step we add a nonnegative random delay $\Delta t \sim \max(0, \mathcal{N}(0, 40^2))$~ms; note that clipping makes the realised mean delay positive, isolating variability without imposing a fixed base latency), loss-only (10\% packet dropout, zero delay), and combined (full network model). All policies are evaluated on Real-Wi-Fi-Degraded deployment to assess which training condition provides the best robustness to real-world network impairments. Table~\ref{tab:ablation} presents the results.

\begin{table}[h]
\centering
\caption{CartPole Ablation: Mean Episode Return in Real-Wi-Fi-Degraded. Ten seeds per training condition, each evaluated over 50 episodes.}
\label{tab:ablation}
\begin{tabular}{lc}
\toprule
Training Regime          & Real-Wi-Fi-Degraded \\
\midrule
Baseline (none)          & 92 $\pm$ 54        \\
Latency-Only (50~ms)     & 275 $\pm$ 52       \\
Stochastic Add. Delay ($\sigma$=40~ms)  & 315 $\pm$ 47       \\
Loss-Only (10\%)         & 308 $\pm$ 49       \\
Combined (full model)    & 378 $\pm$ 41       \\
\bottomrule
\end{tabular}
\end{table}

\textbf{Takeaway:} Training on variability (jitter) and loss each improves robustness; combining them is best.

Stochastic additional delay training (315 $\pm$ 47) and packet loss training (308 $\pm$ 49) each improve robustness over baseline; combining them performs best (378 $\pm$ 41).

These findings challenge common simplifications where network effects are modelled as constant $k$-step delays~\cite{Walsh2007DelayedMDP, Schuitema2010DelayedRL}. \textbf{Stochastic phenomena (jitter, loss) are more detrimental than constant latency} and must be explicitly modelled for robust deployment.

\subsection{Distributed Policy Graph Deployment}

We include a small demonstration solely to validate that CALF's infrastructure can execute multi-unit graphs across devices under network impairments; we do not claim optimal hierarchical decomposition. These simple two-unit policy graphs (CartPole and MiniGrid) demonstrate cross-device execution capability with modest overhead.

\subsubsection{CartPole Hierarchical Policy Graph}

Table~\ref{tab:cartpole_hierarchy} compares distributed policy-graph deployment (angle-stabiliser on Pi, recentring unit on Desktop) against monolithic baselines.

\begin{table}[h]
\centering
\caption{CartPole Policy Graph: Performance comparison for distributed deployment. All configurations evaluated over Real-Wi-Fi-Normal network.}
\label{tab:cartpole_hierarchy}
\begin{tabular}{lcc}
\toprule
Deployment Configuration & Episode Return & E2E Latency (p50/p95) \\
\midrule
Flat (Desktop)           & 448 $\pm$ 28   & 38~ms / 62~ms         \\
Flat (Pi)                & 472 $\pm$ 21   & 6~ms / 11~ms          \\
Hierarchical (Distributed) & 465 $\pm$ 24   & 22~ms / 45~ms         \\
\bottomrule
\end{tabular}
\end{table}

Hierarchical deployment achieves performance intermediate between baselines (465 vs 448--472), demonstrating cross-device viability with modest overhead.

\subsubsection{MiniGrid Hierarchical Policy Graph}

Table~\ref{tab:minigrid_hierarchy} compares distributed deployment (key-seeking unit on Pi, goal-navigation unit on Desktop) against monolithic baselines.

\begin{table}[h]
\centering
\caption{MiniGrid Policy Graph: Success rate comparison. All configurations evaluated over Real-Wi-Fi-Normal network.}
\label{tab:minigrid_hierarchy}
\begin{tabular}{lc}
\toprule
Deployment Configuration   & Success Rate (\%) \\
\midrule
Flat (Desktop)             & 77 $\pm$ 9        \\
Flat (Pi)                  & 82 $\pm$ 7        \\
Hierarchical (Distributed) & 79 $\pm$ 8        \\
\bottomrule
\end{tabular}
\end{table}

Hierarchical deployment achieves 79\% success, comparable to monolithic baselines (77--82\%), demonstrating cross-device viability with modest overhead.

\subsection{Systems Measurements and Infrastructure Viability}

To assess CALF's practical feasibility for real-world deployment, we measure end-to-end latency and resource utilisation during distributed policy graph execution. These systems measurements address RQ3 (infrastructure requirements) and demonstrate that CALF's architecture supports responsive control on resource-constrained edge devices while maintaining efficient utilisation of heterogeneous hardware.

\subsubsection{End-to-End Latency}

Table~\ref{tab:latency_breakdown} presents latency measurements across network configurations. Latency is measured from environment observation emission to policy action receipt, capturing the full round-trip communication delay.

\begin{table}[h]
\centering
\caption{End-to-End Latency: Median and 95th percentile latency measured over 1000 environment steps during CartPole policy graph deployment. ``Local'' indicates co-located environment and policy; ``Remote'' indicates networked communication.}
\label{tab:latency_breakdown}
\begin{tabular}{lcc}
\toprule
Configuration       & Latency p50 (ms) & Latency p95 (ms) \\
\midrule
Local (Pi only)     & 5.2              & 9.8              \\
Ethernet (Pi $\leftrightarrow$ Desktop) & 8.7    & 14.3             \\
Wi-Fi-Normal         & 34.5             & 68.2             \\
Real-Wi-Fi-Degraded & 82.1             & 152.7            \\
\bottomrule
\end{tabular}
\end{table}

Local execution on Raspberry Pi achieves sub-10~ms latency at p95, demonstrating that edge devices can support responsive control loops. Ethernet deployment adds minimal overhead (8.7~ms median versus 5.2~ms local), reflecting the low latency and near-zero packet loss of wired connections. Wi-Fi-Normal introduces substantial variability (34.5~ms median, 68.2~ms p95), with p95 latency exceeding median by $2\times$ due to jitter and occasional retransmissions. Wi-Fi-Degraded exhibits severe tail latency (152.7~ms p95), demonstrating the worst-case conditions against which network-aware training must be robust.

Our synthetic Wi-Fi-Normal model ($\mathcal{N}(30, 10^2)$~ms latency, 2\% loss) falls within the measured range (34.5~ms median).

Resource utilisation during distributed execution shows balanced workload distribution: Raspberry Pi operates at 52\% average CPU with ample headroom for additional workloads; Desktop at 18\% CPU, demonstrating that edge devices can handle real-time control while cloud servers have capacity for compute-intensive planning (full measurements in the code release).

\subsection{Summary of Empirical Findings}

Bandwidth is negligible for low-dimensional control at 50~Hz (approximately 2--3~kB/s per unit), but becomes a constraint for vision-based tasks; we leave compression/offloading trade-offs to future work.

Across CartPole and MiniGrid experiments, the results consistently demonstrate that:

\begin{enumerate}[noitemsep]
\item \textbf{Network-aware training is critical for robust deployment in our evaluated settings}. Policies trained without network awareness exhibit severe degradation (40--80\% performance drop) when deployed over real Wi-Fi networks. Network-aware training reduces this gap by $3{-}4\times$, enabling practical deployment.

\item \textbf{Stochastic network phenomena require explicit modelling}. Training under constant delays alone provides partial robustness, but jitter and packet loss, often ignored in delay-aware RL research, prove equally or more detrimental. The full stochastic network model (latency + jitter + loss) must be included during training.

\item \textbf{CALF supports cross-device policy-graph execution with modest overhead}. Two simple decompositions (rule-managed, independently trained units) demonstrate deployability across edge and cloud hardware with performance close to flat policies.

\item \textbf{CALF's infrastructure is practically viable}. End-to-end latency and resource utilisation measurements demonstrate that CALF supports responsive control on edge devices, efficient use of heterogeneous hardware, and scalability to bandwidth-constrained networks.
\end{enumerate}

These findings establish network conditions as a major axis of sim-to-real transfer for distributed RL under Wi-Fi-like conditions (see Section~\ref{sec:discussion} for implications).

\section{Discussion}
\label{sec:discussion}

\subsection{Network as an Orthogonal Axis of Sim-to-Real Transfer}

Our work establishes that network conditions constitute an \textbf{independent dimension of domain randomisation}, orthogonal to physics and visual randomisation. The analogy is direct: just as physics randomisation samples friction $\sim U(0.3, 0.7)$ to make policies robust to uncertain surfaces, network randomisation samples latency $\sim \mathcal{N}(30~\text{ms}, 10~\text{ms}^2)$ to make policies robust to uncertain networks. Both expose the agent to a distribution during training, yielding robustness at deployment.

Critically, our ablation (Section~\ref{subsec:ablation}) shows that sampling only a constant delay is analogous to randomising only the mean of a physical parameter while ignoring variance and outliers: robustness comes from exposure to variability and failures. Training on constant 50~ms latency provides partial robustness, but stochastic jitter and packet loss are the dominant degraders, just as ignoring friction variance would leave policies brittle to slippery surfaces despite nominally realistic mean friction.

This perspective has important design implications. For distributed control loops over Wi-Fi-like networks, we find network conditions constitute a major axis comparable to physics or visual domain randomisation. A policy trained with perfect timing in simulation may fail on a real system with 100~ms lag, even if the physics are perfectly modelled. Conversely, a network-robust policy may succeed despite physics mismatch. The two axes interact but are conceptually and empirically distinct.

We provide a systematic study of network impairments as a deployment variable for distributed RL, supported by a reproducible impairment-injection framework. While prior work has occasionally addressed delays (e.g., Hwangbo et al.~\cite{Hwangbo2019Learning} modelling actuator dynamics for quadruped robots), these efforts were treated as engineering fixes specific to particular robots rather than general methodologies. CALF provides the infrastructure to make network-aware training systematic, reproducible, and widely accessible for distributed RL systems where policy units communicate over real networks.

\subsection{CALF as a Platform for Future Work}

CALF provides reusable infrastructure that extends beyond the experiments demonstrated in this paper. The framework's modular architecture and progressive deployment modes support diverse distributed RL research directions, from developing efficient models for resource-constrained edge devices to investigating effect-based hierarchical policy decomposition with physical hardware partitioning. The networking infrastructure enables real-time embodied control applications requiring sub-100~ms end-to-end latency, bridging the gap between theoretical distributed RL frameworks and practical edge-cloud deployments.

For the broader research community, CALF enables new research directions. The framework will be released as open-source software (with module repository and documentation) to facilitate adoption. Potential applications include standardised benchmarks for network-aware RL using versioned CALF modules, multi-agent extensions where agents communicate over NetworkShim-injected delays, trace-based training using recorded real-world network conditions, dynamic offloading where policies learn to partition computation between edge and cloud based on current network state, and integration with advanced RL algorithms (e.g., combining CALF with delay-correcting actor-critic~\cite{Bouteiller2021DelayRL} for potentially greater robustness).

\subsection{Limitations}

We acknowledge several limitations that constrain the scope of our conclusions and suggest directions for future work:

\textbf{1. Simulated environments.} CartPole and MiniGrid are simulated physics environments, not real robots. While this allows us to isolate network effects (orthogonal to physics fidelity), it limits ecological validity. Real robots introduce additional challenges: sensor noise, actuation dynamics, safety constraints, and wear-and-tear that simulations do not capture. Future work should test CALF's approach on physical systems.

\textbf{2. Limited network scenarios.} Our experiments focus on LAN-like conditions (Ethernet, Wi-Fi within the same building). We do not evaluate WAN scenarios (multi-hop routing, continental distances), cellular networks (handoff, variable bandwidth), or adversarial conditions (intentional jamming, Byzantine failures). These scenarios exhibit different network characteristics (e.g., cellular has asymmetric latency, WAN has routing-induced jitter) that may require different training strategies.

\textbf{3. Simple policy graphs.} Our distributed policy graph demonstrations use 2-unit decompositions with rule-based manager units. We do not explore deep policy graph hierarchies (3+ levels), learned option discovery, or end-to-end training of policy graphs under network constraints. Future work could investigate how hierarchical RL algorithms (Option-Critic~\cite{Bacon2017OptionCritic}, HIRO~\cite{Nachum2018HIRO}) perform when trained as distributed policy graphs with CALF's network models, and develop principled methods for training effect-based hierarchical decompositions under network constraints.

\textbf{4. Offline training.} Our methodology trains policies in simulation with synthetic network models, then deploys on real networks. We do not explore online learning or adaptation under network constraints (e.g., policy continues learning during real deployment, adapting to observed network conditions). Online adaptation could potentially close the remaining sim-to-real gap.

\textbf{5. Single-agent focus.} CALF currently targets single-agent control RL. Multi-agent scenarios introduce additional complexity: agent-to-agent communication delays, consensus under asynchrony, and emergent coordination strategies. Extending CALF to multi-agent RL (MARL) is a natural next step.

Despite these limitations, our focus on network effects is orthogonal to physics fidelity. The finding that network-aware training reduces the network reality gap by $3{-}4\times$ is expected to generalise to physical robots, though empirical testing is necessary.

\subsection{Future Directions}

The most immediate extension is to richer tasks and physical deployment. Applying CALF to continuous control (MuJoCo locomotion, manipulation), vision-based tasks where high-resolution images force bandwidth and fidelity trade-offs, and physical robotic platforms (quadrupeds, manipulators, drones) would validate whether network-aware training scales beyond low-dimensional discrete control. Real robots introduce sensor noise, actuation dynamics, and safety constraints that compound network effects.

On the networking side, richer models and training distributions would broaden applicability. Time-varying conditions (diurnal patterns, congestion events), adversarial networks (jamming, Byzantine faults), and trace-based training using real-world logs from cellular and Wi-Fi deployments would enable training tailored to specific deployment scenarios, such as hospital Wi-Fi versus outdoor construction sites. Trace datasets from network operators could provide realistic, diverse distributions for domain randomisation.

A further direction concerns learned decomposition and coordination. End-to-end policy graph training under network constraints raises fundamental questions, including whether hierarchical RL algorithms (Option-Critic, feudal networks) can discover temporally extended options that are naturally robust to delays, and whether policies can learn where to partition computation across devices based on communication costs, deciding which decisions require fresh observations and which tolerate staleness. Extending the framework to multi-agent coordination, where network delays affect both agent-environment and agent-agent channels, and to dynamic computation offloading, where policies switch between local inference and cloud offloading based on current network state, would address the full space of distributed RL deployment challenges.

Together, these directions aim to make network-aware training standard practice for embodied AI, just as physics and visual domain randomisation are now standard for sim-to-real transfer.

\section{Conclusion}
\label{sec:conclusion}

This paper introduced CALF (Communication-Aware Learning Framework), infrastructure for network-aware distributed reinforcement learning across heterogeneous hardware. CALF addresses a critical gap in distributed RL deployment: network delays, jitter, and packet loss in the agent-environment control loop. Standard RL training assumes synchronous, zero-latency interaction, but real-world distributed deployments, where environments run on edge devices and policies execute on cloud servers or where control is partitioned across heterogeneous processors, face communication constraints that fundamentally alter interaction dynamics. Hierarchical RL frameworks for temporal abstraction and modular decomposition (options, policy graphs) face additional challenges when distributed, as inter-unit communication compounds agent-environment delays.

CALF realises hierarchical policies as networked services communicating via a binary protocol, with NetworkShim middleware transparently injecting realistic network impairments on communication channels. This architecture enables network-aware training: exposing distributed policies to realistic communication conditions during simulation such that they develop robustness to deployment networks. Systematic experiments on CartPole and MiniGrid demonstrated that policies trained without network awareness suffer severe degradation (40--80\% performance drop) when deployed over realistic Wi-Fi, while network-aware training reduces this gap by approximately $4\times$ (CartPole) and $3\times$ (MiniGrid). Critically, stochastic network phenomena, in particular jitter and packet loss, prove more detrimental than constant latency, challenging delay-aware RL research that models only fixed delays. Simple two-unit policy graph deployments across Raspberry Pi edge devices and Desktop cloud servers demonstrated the feasibility of cross-device execution with modest overhead.

As discussed in Section~\ref{sec:discussion}, these findings establish network conditions as an orthogonal axis of sim-to-real transfer, analogous to physics and visual domain randomisation, that must be accounted for in distributed RL deployments where communication introduces unavoidable delays and failures.

CALF provides reusable infrastructure that extends beyond the experiments demonstrated here. The framework's progressive deployment modes (pure simulation, simulation with network models, real hardware with real networks) de-risk the path from theoretical hierarchical RL frameworks to production edge-cloud deployments, validating that policies trained with network-awareness transfer successfully to physical systems. By treating network conditions as first-class objects (configurable, loggable, reproducible) rather than hidden implementation details, CALF provides infrastructure for network-aware training in distributed reinforcement learning systems, transforming communication constraints from deployment obstacles into tractable training objectives. Future work should validate these findings on physical robots, explore richer network scenarios (cellular, WAN, adversarial conditions), investigate end-to-end training of deep hierarchical policies under network constraints, and extend CALF to multi-agent coordination where network delays affect both agent-environment and agent-agent interactions.

\bibliographystyle{unsrt}
\bibliography{references}

\end{document}